\documentclass{article}

\usepackage[final]{nips_2016}
\usepackage{tabularx} % extra features for tabular environment
\usepackage{mathtools}  % improve math presentation
\usepackage{cases}
\usepackage{subcaption} % for multiple figures
\usepackage{sidecap} % for side captions
\usepackage[utf8]{inputenc} % allow utf-8 input
\usepackage[T1]{fontenc}    % use 8-bit T1 fonts
\usepackage{hyperref}       % hyperlinks
\usepackage{url}            % simple URL typesetting
\usepackage{booktabs}       % professional-quality tables
\usepackage{amsfonts}       % blackboard math symbols
\usepackage{nicefrac}       % compact symbols for 1/2, etc.
\usepackage{microtype}      % microtypography
\usepackage{color}

\title{Theory and Tools for the Conversion of Analog to Spiking Convolutional Neural Networks}

\author{
  Bodo Rueckauer, Iulia-Alexandra Lungu, Yuhuang Hu, and Michael Pfeiffer \\
  Institute of Neuroinformatics \\
  University of Zurich and ETH Zurich \\
  Winterthurerstrasse 190 \\
  8057 Zurich, Switzerland \\
  \texttt{\{rbodo,iulialexandra,yuhu,pfeiffer\}@ini.uzh.ch} 
}

\begin{document}

\maketitle

\begin{abstract}

Deep convolutional neural networks (CNNs) have shown great potential for numerous real-world machine learning applications, but performing inference in large CNNs in real-time remains a challenge. We have previously demonstrated that traditional CNNs can be converted into deep spiking neural networks (SNNs), which exhibit similar accuracy while reducing both latency and computational load as a consequence of their data-driven, event-based style of computing. Here we provide a novel theory that explains why this conversion is successful, and derive from it several new tools to convert a larger and more powerful class of deep networks into SNNs. We identify the main sources of approximation errors in previous conversion methods, and propose simple mechanisms to fix these issues. Furthermore, we develop spiking implementations of common CNN operations such as max-pooling, softmax, and batch-normalization, which allow almost loss-less conversion of arbitrary CNN architectures into the spiking domain. Empirical evaluation of different network architectures on the MNIST and CIFAR10 benchmarks leads to the best SNN results reported to date.
\end{abstract}

\section{Introduction}

Computation in Spiking Neural Networks (SNNs) is event-based and data-driven, and thus neurons only update whenever new relevant information needs to be processed. This contrasts with the frame-based computations performed by traditional Analog Neural Networks (ANNs), which process the entirety of the input, followed by computing all neuronal activations layer-by-layer before producing a final output. Recent results such as \citep{neil2016learning,farabet2012comparison,oconnor2013real,zambrano2016fast} have shown that the event-based mode of operation in SNNs is particularly attractive for reducing latency and computational load in deep neural networks, which represent the state of the art in most machine learning benchmarks \citep{lecun2015deep}. Deep SNNs can be queried for results already after the first output spike is produced, unlike ANNs where the result is available only after all layers have been completely processed \citep{Diehl2015Fast}. 
SNNs are also naturally suited to process input from event-based sensors \citep{liu2014event,posch2014retinomorphic}, but even for classical frame-based machine vision applications such as object recognition or detection, SNNs have been shown to be accurate, fast, and efficient, in particular when run on neuromorphic hardware platforms \citep{Esser20092016,Neil2013Minitaur,stromatias2015scalable}. SNNs could thus play an important role in supporting, or in some cases replacing deep ANNs in tasks where fast and efficient classification in real-time is crucial, such as detection of objects in larger and moving scenes, tracking tasks, or activity recognition \citep{hu2016dvs}.
%Conventional tasks such as face detection and object recognition can benefit from the low latency and reduced number of operations performed in event-based networks even when the input is not event-based, especially when the inference is performed on dedicated hardware that accelerates spiking neural networks, such as novel neuromorphic platforms. 

%\MP{First talk about deep networks and why they are successful, then the conversion to SNNs}

Training Deep SNNs directly from their spiking activity is notoriously difficult, and only recently methods for backpropagation-like training of SNNs have been developed \citep{lee2016training}. However, a number of studies have shown that SNNs can be successfully constructed by converting conventionally trained ANNs, relating the activations of ANN units to firing rates of spiking neurons. 
The purpose of this study is to identify a number of challenges in the conversion of networks through a novel theory, and proposing new mechanisms which significantly improve the performance of deep SNNs. 

Previous work on ANN-to-SNN conversion starts with \citep{Perez-Carrasco2013}, where CNN units were translated into biologically inspired spiking units with leaks and refractory periods, aiming for processing of inputs from event-based sensors. \citep{cao2015spiking} suggested a close link between the transfer function of a spiking neuron, i.e. the relation between input current and output firing frequency to the activation of a rectified linear unit (ReLU), which is nowadays the standard model for units in ANNs. They achieved good performance on conventional computer vision benchmarks, converting a class of CNNs that was restricted to having zero bias and only average-pooling layers. Their method was improved by \citep{Diehl2015Fast}, who achieved nearly loss-less conversion of ANNs for the MNIST task through \emph{weight normalization}. This technique rescales the weights to avoid approximation errors in SNNs due to either excessive or too little firing. \citep{Hunsberger2015Spiking} introduced a conversion method where noise injection during training improves the robustness to approximation errors of the SNN with biologically more realistic neuron models. 
\citep{Esser20092016} demonstrated an approach that optimized CNNs for the neuromorphic TrueNorth platform with low-precision weights and restricted connectivity.
%proposed a mapping of conventional CNNs onto SNNs amenable to hardware implementation. Their method suffers from limitations imposed by the neuromorphic platform they are using, such as binary activations, trinary weights, the inability to implement biases as well as wasteful duplications of multiple units. The best performance they report on CIFAR10 is 87.50\%, using 8 TrueNorth chips. While their result is marginally lower than the highest accuracy we obtained, their method demands notably more computational resources. 
Recently, \citep{zambrano2016fast} have developed adapting SNNs, which encode information with a minimum number of spikes, thereby achieving good accuracy with an order of magnitude less spikes than in other SNN approaches.
All these approaches showed that SNNs can achieve near state-of-the-art accuracy, while at the same time improving classification speed, and aiming towards constructing deep SNNs that can run on low-power neuromorphic platforms.

%For this they developed an adapting SNN, where individual neurons modify their threshold, reset value and size of their action potential in order to optimize their firing rate online. This approach reduces the number of spikes by an order of magnitude without impairing classification performance on MNIST. It is an open question how well the adapting SNN scales to larger networks, where the increased variance due to low spike-rates tends to have a devastating effect on classification accuracy.

% In previous work such as \citep{cao2015spiking,Diehl2015Fast} a heuristic was suggested to link the transfer function of a spiking neuron, i.e. the relation between input current and output firing frequency to the activation of a rectified linear unit (ReLU), which is nowadays the standard model for units in ANNs. Indeed, there is a close resemblance between the ReLU activation function $y(x) = \max(0, x)$ with transfer functions of typical simple spiking neuron models such as the (leaky) integrate-and-fire neuron \citep{gerstner2014neuronal}. Such conversion algorithms start from an analog neural network trained using standard machine learning methods, such as backpropagation. The ANN weights are applied to synaptic connections within a functionally equivalent SNN. The goal of converting analog to spiking networks is to provide deep SNN models with near state-of-the-art accuracy, while at the same time achieving the potential advantages of event-based approaches in terms of speed, reduction of computation, and lower energy consumption for reaching this desired level of accuracy.

In this work we investigate in detail the conversion algorithm proposed in \citep{Diehl2015Fast}, which can convert small ANN models into SNNs with minimal accuracy loss (less than 1\% on MNIST). However, we found that running the same algorithms without modifications on larger networks (e.g. for CIFAR10) leads to an unacceptable drop in accuracy on the order of 10\% or more. In our tests we identified that the main reason for a drop of performance is a reduction of firing rates in higher layers, which can arise due to overly pessimistic weight normalization. Furthermore, the algorithms proposed by \citep{Diehl2015Fast} and \citep{cao2015spiking} had severe restrictions on what kind of CNNs could be converted. This prevented using features commonly used in the most successful CNNs for visual classification, such as max-pooling, softmax, batch normalization \citep{Ioffe2015}, or even basic features of neuron models such as biases.  In the following we first analytically investigate the problem of ANN-to-SNN conversion in Section \ref{sec:theory}, before introducing a set of new tools and tricks that expand the class of networks that can be converted and improve accuracy (Section \ref{sec:methods}). The contribution of the proposed methods is evaluated on the CIFAR10 benchmark in Section \ref{sec:results}.
 
%\MP{Probably shorten this}
%\citep{Hunsberger2015Spiking} introduced a conversion method using biologically plausible Leaky Integrate and Fire (LIF) neurons for the SNN, as well as rate-based versions of the Leaky Integrate and Fire (LIF) model for the ANN. They suggest noise injection during training as a means of improving SNN robustness to approximation errors. Similar to \citep{Diehl2015Fast}, they replace max-pooling layers by average pooling. Furthermore, they do not perform weight normalization. The accuracy they obtain on MNIST and CIFAR10 is 98.37\% and 82.95\% respectively, which is below the results reported in our study. 

%Another study employing more biologically plausible neuron models is the one by \citep{Perez-Carrasco2013}. Their work illustrates the advantage of SNNs coupled with event-based sensors in terms of speed. However, as their method was aimed for event-based inputs, it has not been tested on standard machine learning benchmark datasets. 

\section{Theory of Conversion of ANNs into SNNs}
\label{sec:theory}

In this section we investigate analytically how firing rates in SNNs approximate ReLU activations in ANNs. This was suggested first by \citep{cao2015spiking} as the basis of ANN-to-SNN conversion, but  a theoretical basis for this principle so far has been lacking. From the basic approximation equations we derive a simple modification of the reset mechanism after spikes, which turns each SNN neuron into an unbiased approximator of the target function.
We assume here a one-to-one correspondence between ANN units and SNN neurons, even though it is also possible to represent each ANN unit by a population of spiking neurons. For a network with $L$ layers let $\mathbf{W}^l, l \in \{1, \ldots, L\}$ denote the weight matrix connecting units in layer $l-1$ to layer $l$, with biases  $\mathbf{b}^l$. The number of units in each layer is $M^{l}$. In the ANN, ReLU activations of neuron $i$ in layer $l$ are computed according to
\begin{equation}
a^l_i := \max\left(0, \sum^{M^{l-1}}_{j=1}W^l_{ij}a^{l-1}_j + b^l_i\right),
\label{eq:ann_activ}
\end{equation}
starting with $\mathbf{a}^0 = \mathbf{x}$ as the input, which is assumed to be normalized so that each $x_i \in [0, 1]$. 

Each SNN neuron has a membrane potential $V^l_i(t)$, which is driven by the input current
\begin{equation}
z^l_i(t) := \tau\left(\sum^{M^{l-1}}_{j=1}W^l_{ij}\Theta^{l-1}_{t,j} + b^l_i\right) ~~~,
\label{eq:SNNinput}
\end{equation}
where $\Theta_{t,i}^l$ is a step function indicating the occurrence of a spike at time $t$:
\begin{equation}
\Theta^l_{t,i} := \Theta(V^l_i(t-1) + z^l_i(t) - \tau), \; \text{with } \Theta(x) = 
  \begin{cases}
    1 & \quad \text{if } x \geq 0 \\
    0 & \quad \text{else.} \\
  \end{cases}
  \label{eq:stepfunc}
\end{equation}

Every input pattern is presented for $T$ time steps, with time step $dt \in \mathbb{R}^+$. The highest firing rate supported by the simulator is given by the inverse time resolution $r_\mathrm{max} := 1/dt$, and input rates are proportional to the pixel intensity or RGB value. We can compute the firing rate of each SNN neuron as $r^l_i(T) := N^l_i(T) / T$, where  $N^l_i(T) := \sum^T_{t=1}\Theta^l_{t,i}$ is the number of spikes generated.
The principle of ANN-to-SNN conversion, as introduced in \citep{cao2015spiking,Diehl2015Fast} is that the firing rates $r^l_i$ should correlate with the original ANN activations $a_i^l$ in (\ref{eq:ann_activ}) such that $r_i^l(T)\rightarrow a_i^l r_\mathrm{max}$. 
This relationship is formalized by introducing a membrane equation for each spiking neuron, and estimating the mean firing rates $r^l_i(T)$ in a next step.
%In order to derive a relation between these two quantities, we first write down the membrane equation of a spiking neuron, and in a second step perform a time-average over the membrane equation to obtain mean firing rates $r^l_i(T)$.

\paragraph{Membrane equation}
The spiking neuron integrates inputs $z^l_i(t)$ until the membrane potential $V^l_i(t-1)$ exceeds a threshold $\tau \in \mathbb{R}^+$ and a spike is generated. At that time, the membrane potential is reset, and here we compare two types of reset: \emph{reset to zero}, which is used in \citep{Diehl2015Fast} always sets the membrane potential back to a baseline, typically zero, whereas \emph{reset by subtraction} subtracts the threshold $\tau$ from the membrane potential at the time where the threshold is exceeded:
\begin{subnumcases}{V^l_i(t)=}
  \left(V^l_i(t-1) + z_i^l(t)\right)\left(1-\Theta^l_{t-1, i}\right) & reset to zero \label{eq:membrane_equ1} \\
  \ \ V^l_i(t-1) + z^l_i(t) \quad - \tau\Theta^l_{t-1, i} & reset by subtraction \label{eq:membrane_equ2}
\end{subnumcases}

From these membrane equations we can derive slightly different approximation properties for the two reset mechanisms. For simplicity we first assume that the input currents $z^1_i = \tau a_i^1 > 0$ remain constant over time, that $V_i^l(0)=0$, and analyze only the first hidden layer. In the \emph{reset to zero} case this implies that there will always be a constant number of time steps $n$ between spikes of the same neuron, and the threshold will always be exceeded by the same constant amount $\epsilon_i^1 = V_i^1(n) - \tau = n\cdot z_i^1 - \tau$. The firing rate of this neuron $r_i^1(t)$ will therefore be $r_{\max}/n$, but one can easily derive that in an exact non-time stepped simulation, the threshold would have been crossed earlier, namely at $t^*=n \cdot \frac{\tau}{\tau+\epsilon_i^1}$. Given that $\tau+\epsilon_i^1 = V_i^1(n) = n\cdot \tau\cdot a_i^1$ we see that $r_\mathrm{max} / t^*$ would be the correct estimate of the firing rate, and the reset to zero mechanism inevitably leads to an approximation error, which appears as the second term on the right in
\begin{equation}
r_i^1(t) = a_i^1 r_\mathrm{max} - r_\mathrm{max} \frac{\epsilon_i^1}{n\cdot \tau}.
\label{eq:reset0error}
\end{equation}

This error depends mainly on $\epsilon_i^1$ and does not go away simply with longer simulation time. For shallow networks and easy tasks such as MNIST this error seems to be a minor problem, but we have found that an accumulation of approximation errors in deeper layers degrades the accuracy. We also see from (\ref{eq:reset0error}) that larger $\tau$ and smaller inputs improve the approximation, at the expense of longer integration times. This is because $\epsilon_i^1$ is bounded by $z_i^1$ if $n>1$, and also because $n$ will increase for smaller inputs. This is further explanation why the weight normalization scheme proposed in \citep{Diehl2015Fast}, according to which network activations never exceed unity improves performance in the reset-to-zero case. Another obvious possibility to improve the approximation is to reduce the simulation time step, but this comes at the cost of increased computational effort.

%\BR{It seems a bit unfair to include the sampling error V(T)/T in the reset-by-subtraction but not in the reset-to-zero...}
%\MP{In the reset-to-zero case it matters much less, because it goes away with time but the other error stays constant}

A simple switch to the \emph{reset by subtraction} mechanism improves the approximation, and makes the conversion scheme suitable also for deeper networks. In this case the time between spikes is not constant, and the membrane potential at any time point is given as $V_i^1(T) = T \cdot z_i^1 - N_i^1(T) \cdot \tau$. From this we get $N_i^1(T) = \left \lfloor \frac{T\cdot z_i^1-V_i^1(T)}{\tau} \right \rfloor$, and can estimate 
\begin{equation}
r_i^1(T) = \frac{N_i^1(T)}{T}=\frac{z_i^1}{\tau}-\frac{V_i^1(T)}{\tau T} = a_i^1 r_\mathrm{max} - \frac{1}{\tau T}V_i^1(T).
\label{eq:eq:subtract_approx}
\end{equation}
This means that the firing rate estimate converges to its target value $a_i^1 \cdot r_\mathrm{max}$, with the only approximation error due to the discrete sampling. This mechanism therefore leads to more accurate approximations of the underlying ANN than in the methods proposed in \citep{cao2015spiking,Diehl2015Fast}. We will show later in Section \ref{sec:results} that this results in improved accuracy in larger networks.
A potential problem exists in the case where $z_i^1 > \tau$. In this case the best a neuron can do is to fire with its maximal firing rate $r_\mathrm{max}$, but can never fully reach its target frequency. This again makes the case for using weight normalization as in \citep{Diehl2015Fast}, in order to prevent saturation of firing.

\paragraph{Firing rates in higher layers}
The previous results were based on the assumption that the neuron receives a constant input $z$ at each step of the simulation. When using spiking neurons in the hidden neurons, this condition only holds for the first hidden layer and input in the form of analog currents instead of irregular spike trains.
%, where the input does not depend on the dynamics of the spike generation mechanism. 
For the \emph{reset-by-subtraction} case we can analytically derive how the approximation error propagates through the deeper layers of the network. For this we insert the expression for SNN input $z^l_i$ from (\ref{eq:SNNinput}) into the membrane equation \eqref{eq:membrane_equ2} for $l>1$, average $V_i^l(t)$ over the simulation time $T$, and solve for the firing rate $r^l_i(T)$. This yields
%by 1. inserting our expression for the SNN input  2. averaging over the whole simulation time (applying $1/T\sum^T_{t=1}$), and 3. solving for the spikerate $r^l_i$. This procedure yields
\begin{equation}
r^l_i(T) = \sum^{M^{l-1}}_{j=1}W^l_{ij}r^{l-1}_j(T) + r_\mathrm{max} b^l_i - \frac{V^l_i(T)}{\tau T}.
\label{eq:r2b}
\end{equation}
\noindent
This result is not surprising, since the firing rate of a neuron in layer $l$ is given by the weighted sum of the firing rates of the previous layer, minus the time-decaying approximation error that was also found in the first layer. This shows that the approximation errors of earlier layers are propagated through the network, and multiplied with the weights of the next higher layer. The recursive expression \eqref{eq:r2b} can be solved iteratively by inserting the rates of the previous layer rates, starting with the known rates of the first layer \eqref{eq:eq:subtract_approx}:
\begin{equation}
r^l_i = a^l_i r_\mathrm{max} - \Delta V^l_{i_l}- \sum^{M^{l-1}}_{i_{l-1}=1}W^l_{i_l i_{l-1}}\Delta V^{l-1}_{i_{l-1}} - \cdots - \sum^{M^{l-1}}_{i_{l-1}=1}W^l_{i_l i_{l-1}}\cdots\sum^{M^{1}}_{i_1=1}W^2_{i_2 i_1}\Delta V^{1}_{i_{1}}
\end{equation}
\noindent
with $\Delta V^l_i := V^l_i(T) / (\tau T)$. Thus, a neuron $i$ in layer $l$ receives an input spike train with a slightly lower spike rate, reduced according to the sampling error $\Delta V$ of previous layer neurons. These errors accumulate for higher layers, which explains why it takes longer to achieve high correlations of ANN activations, and why SNN firing rates deteriorate in higher layers.
%\MP{But they should disappear over time, right?}
%\BR{Yes. In fact, it seems to me now that the only advantage of 99.9th percentile normalization over max norm is the speed of reaching the optimal accuracy. A lower percentile simply increases firing rates and allows faster guessing. But a network normalized by the max will eventually also reach a high performance, namely when the sampling error vanishes in the limit of large $T$. I just ran CIFAR for 700 ms instead of 300 ms and got 87.13\% (before: 83.60\%), which is only slightly below our 99.9th percentile result (87.32\% after 140 ms). On the other hand, 99.9th percentile norm has the disadvantage of reducing ANN-SNN-correlation due to saturation. So this again boils down to a speed-accuracy trade-off. I guess we cannot sell the percentile-method just by talking about improved accuracy; we need to bring speed into the discussion.}

\section{New Methods for ANN-to-SNN Conversion}
\label{sec:methods}
In the following we introduce new methods and heuristics that improve the classification accuracy of deep SNNs, by either allowing the conversion of a wider ranger of ANNs, or by reducing approximation errors in the SNN.

\subsection{Converting biases}
Biases are standard in ANNs, but were explicitly excluded by previous conversion methods for SNNs. In a spiking network, a bias can simply be implemented with a constant input current (proportional to the ANN bias) to each cell's membrane potential. The theory in Section \ref{sec:theory} fully applies to the case of neurons with biases, and the following Section \ref{sec:normalization} shows how parameter normalization can be applied to biases as well.

\subsection{Parameter normalization}
\label{sec:normalization}
One source of approximation errors is that in an SNN in time-stepped simulation neurons are restricted to a firing rate range of $[0, r_{\mathrm{max}}]$, whereas ANNs have no such constraints. \citep{Diehl2015Fast} have introduced weight normalization as a means to avoid approximation errors due to too low or too high firing, thereby significantly improving the performance of converted SNNs.
%by maintaining high spike rates while avoiding saturation. Normalization is not necessary for layers without parameters, such as pooling or flattening layers.
Here we extend the \emph{data-based weight normalization} mechanism introduced in \citep{Diehl2015Fast} to the case of neurons with biases and suggest a heuristic that makes the normalization process more robust to outliers.
%Furthermore, we present a more robust normalization method, which is less aggressive in reducing weights, thereby leading to higher firing rates and therefore better approximation of the ANN. Variations of this method using layer-wise normalization, online normalization, and dealing with extremely active neurons are also presented.

\subsubsection{Normalization with biases}
\label{sec:normbias}
The \emph{data-based normalization} scheme from \citep{Diehl2015Fast} is based on the linearity of the ReLU unit used for ANNs. It can simply be extended to biases by linearly rescaling all weights such that the ANN activation $a$ is smaller than $1$ for all training examples. In order to preserve the information encoded within a layer, the parameters of a layer need to be scaled jointly. Denoting the maximum ReLU activation in layer $l$ as $\lambda^{l}=\max[\mathbf{a}^{l}]$, then weights $\mathbf{W}^l$ and biases $\mathbf{b}^l$ are normalized to $\mathbf{\tilde W}^l \leftarrow \mathbf{W}^l \frac{\lambda^{l-1}}{\lambda^l}$ and $\mathbf{\tilde b}^l \leftarrow \mathbf{b}^l / \lambda^l$.
%We refer to \citep{Diehl2015Fast} for pseudo-code, and note that biases can simply be scaled by the inverse of the maximum activation in the layer, otherwise no changes are necessary.
%\begin{enumerate}
%\item Normalize the input $\mathbf{x} = \mathbf{a}^0$: $\; \mathbf{\tilde a}^0 = \mathbf{a}^0/\lambda^0$, with $\lambda^0 = \max[\mathbf{a}^0]$
%\item For all layers $l\in [1, ..., L]$ do
%  \begin{enumerate}
%  \item Compute the ANN activations
%  $\mathbf{a}^l=f(\lambda^{l-1}\mathbf{W}^l \mathbf{\tilde a}^{l-1} + \mathbf{b}^l])$ of layer $l$, where $f(x)=\max[0,x]$ is the ReLU activation function and the scale $\lambda^{l-1}=\max[\mathbf{a}^{l-1}]$ the maximum activation of layer $l-1$.
% \item Normalize parameters $[\mathbf{\tilde W}^l, \mathbf{\tilde b}^l] \leftarrow [\mathbf{W}^l \lambda^{l-1}/\lambda^l, \mathbf{b}^l/\lambda^l]$.
%  \item Replace the old parameters:
%  $[\mathbf{W}^l, \mathbf{b}^l] \leftarrow [\mathbf{\tilde W}^l, \mathbf{\tilde b}^l]$.
%  \end{enumerate}
%\end{enumerate}
%With this procedure, all activations in the ANN become smaller than one, and the scale can simply be absorbed in the layer parameters: $\mathbf{\tilde a}^l = \mathbf{a}^l / \lambda^l = f(\lambda^{l-1}/\lambda^l\mathbf{W}^l\mathbf{\tilde a}^{l-1} + \mathbf{b}^l/\lambda^l]) = f(\mathbf{\tilde W}^l \mathbf{a}^{l-1} + \mathbf{\tilde b}^l])$. This scheme therefore prevents SNN neurons from reaching the saturation regime after conversion.

\subsubsection{Robust mormalization}
\begin{figure}[htbp]
\centering
\begin{subfigure}{.4\textwidth}
\centering
\includegraphics[scale=0.3]{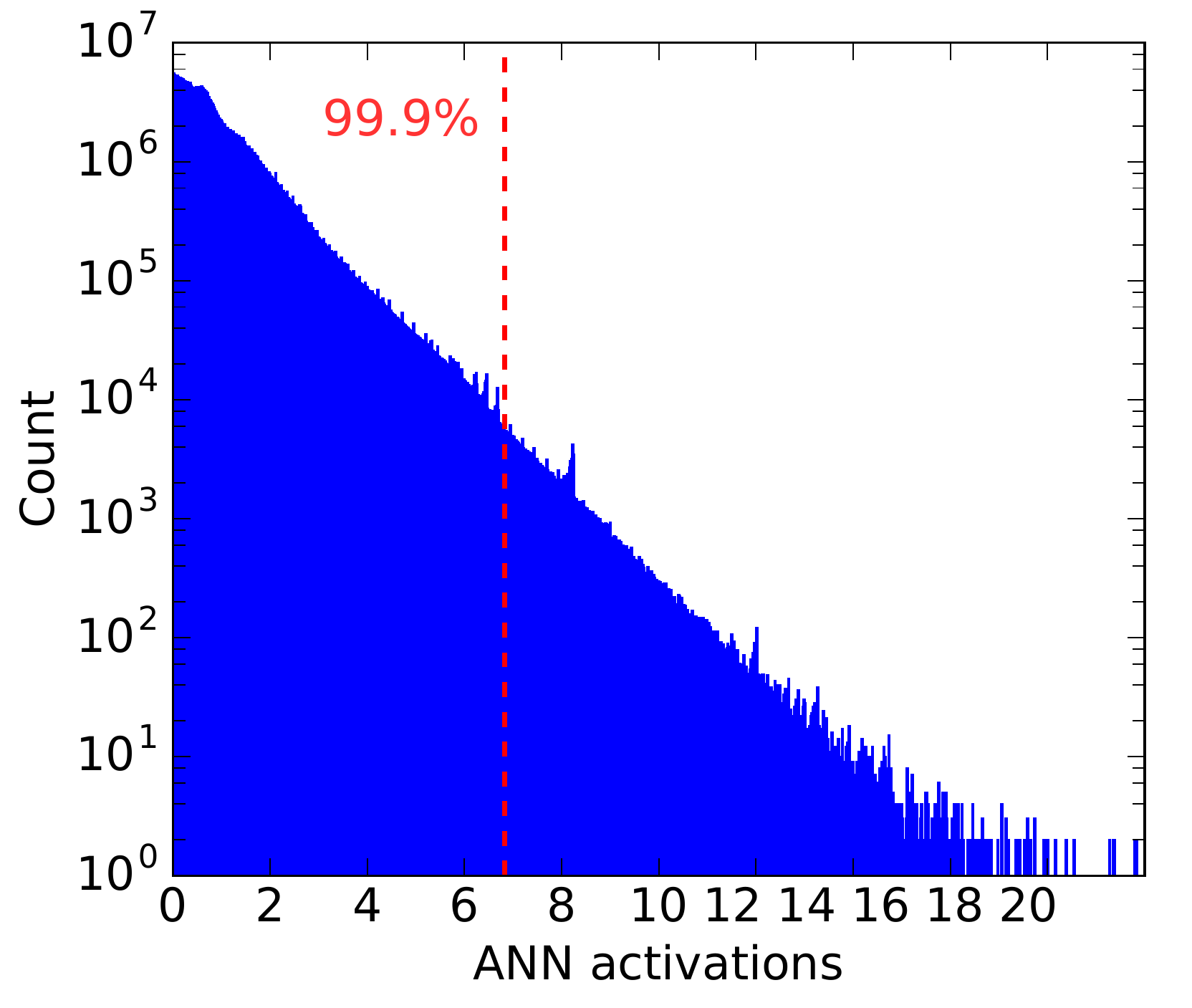}
\caption{}
\label{fig:activations}
\end{subfigure}%
\centering
\begin{subfigure}{.4\textwidth}
\includegraphics[scale=0.3]{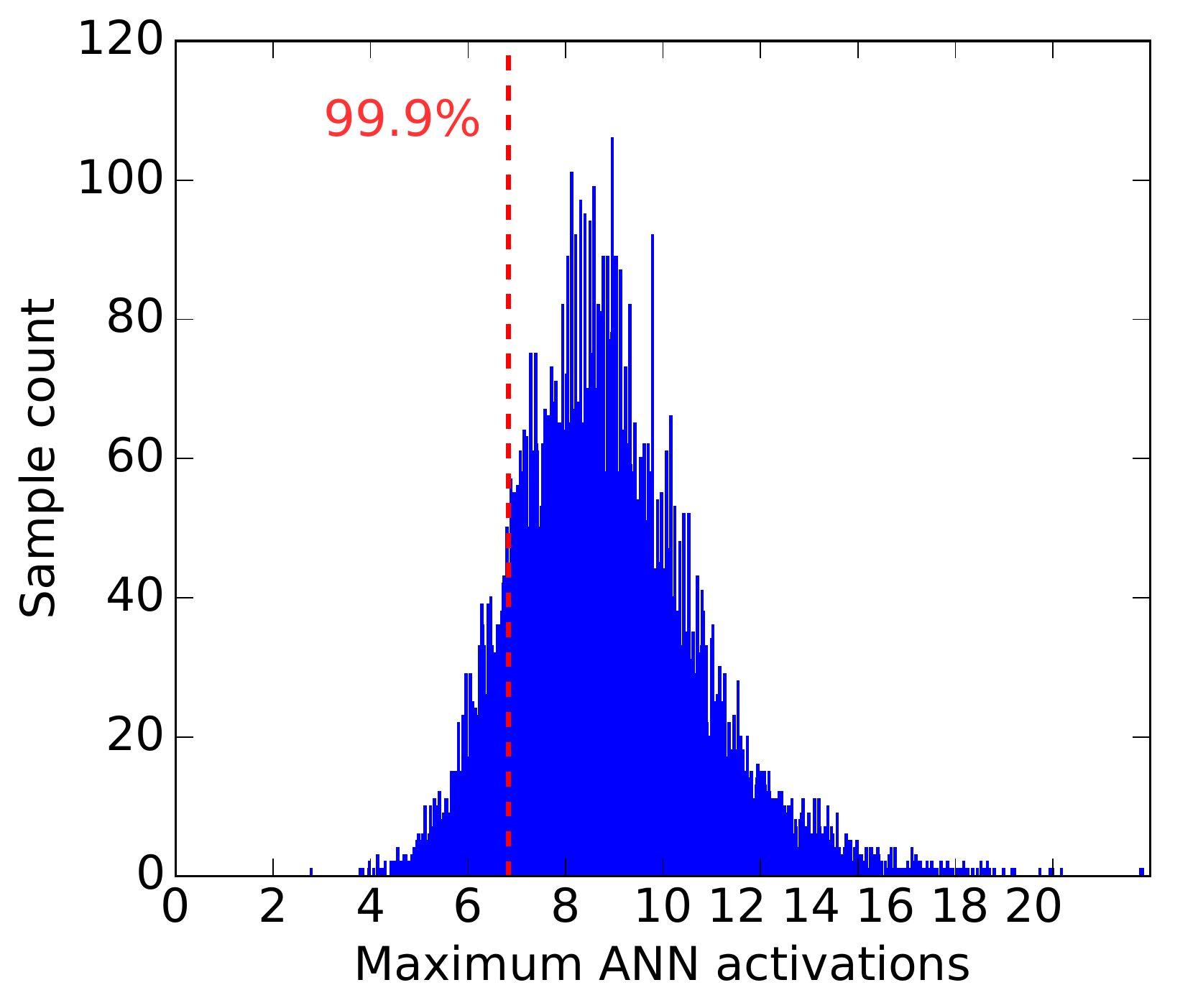}
\caption{}
\label{fig:max_activations}
\end{subfigure}
\caption{(a) Distribution of all non-zero activations in the first convolution layer of a CNN, for 16666 CIFAR10 samples, in log-scale. The dashed line in both plots indicates the 99.9th percentile of all ReLU activations across the dataset, corresponding to a normalization scale $\lambda=6.83$. This is more than three times less than the overall maximum of $\lambda_{max}=23.16$. (b) Distribution of maximum ReLU activations for the same 16666 CIFAR10 samples. For most samples their maximum activation is far from $\lambda_{max}$.}
\end{figure}

Although weight normalization avoids saturating firing rates in the SNN, it might result in very low firing rates, thereby increasing the latency until information reaches the higher layers. 
%Avoiding saturation might lead to another undesired effect in the SNN, namely too low firing rates, which causes insufficient information to propagate to higher layers. 
In the algorithm sketched above in Section \ref{sec:normbias}, which we refer to as "max-norm", the normalization factor $\lambda^l$ by which the weights and biases are scaled was set to the maximum ANN activation among all samples of the training set. This is a very conservative approach, which ensures that the SNN firing rates never exceed the maximum firing rate. The drawback is that this procedure is prone to be influenced by singular outlier samples that lead to very high activations, while for the majority of the remaining samples, the firing rates will remain considerably below saturation. 
%Approximation errors are amplified in the low firing-rate regime, so this "max-norm" normalization might reduce the accuracy for many samples. 
Such outliers are not uncommon, as shown in Figure \ref{fig:activations}, which plots the distribution of all non-zero activations in the first convolution layer for 16666 CIFAR10 samples (in log-scale). The maximum observed activation is more than three times higher than the 99.9th percentile. Figure \ref{fig:max_activations} shows the distribution of the highest activations across the 16666 samples for all ANN units in the same layer, revealing a large variance across the dataset, and a peak that is far away from the absolute maximum. This explains why normalizing by the maximum can potentially perform poorly: For the vast majority of samples even the maximum activation of units within a layer will lie far below the chosen normalization scale, and thus there is insufficient firing in the SNN to drive higher layers and obtain accurate results. 

As a more robust alternative we propose that instead of choosing the maximum activation, we can set $\lambda$ to the $p$-th percentile of the total activity distribution. This discards extreme outliers, and increases SNN firing rates for a larger fraction of samples. The potential drawback is that a small percentage of neurons will saturate, so choosing the normalization scale involves a trade-off between saturation and insufficient firing. In the following, we refer to the percentile $p$ as the "normalization scale", and note that the "max-norm" method is recovered as the special case $p=100$. Typical values for $p$ that perform well are in the range $[99.0, 99.999]$. In general, saturation of a small fraction of neurons seems to degrade network performance less than having too low spike rates. This method can be combined with using batch-normalization during ANN training \citep{Ioffe2015}, which standardizes the activations in each layer and therefore produces fewer extreme outliers.

\subsection{Conversion of Batch-normalization layers}
Batch-normalization (BN) \citep{Ioffe2015} reduces internal covariate shift in ANNs and thereby speeds up the training process. BN introduces additional layers where affine transformations of inputs are performed in order to achieve zero-mean and unit variance. An input $x$ is transformed into $\mathrm{BN}[x]= \frac{\gamma}{\sigma}(x-\mu)+\beta$, where mean $\mu$, variance $\sigma$, and the two learned parameters $\beta$ and $\gamma$ are all obtained during training as described in \citep{Ioffe2015}. After training, these transformations can be integrated into the weight vectors, thereby preserving the effect of BN, but eliminating the computations. Specifically, we set $\tilde W^l_{ij} = \frac{\gamma^l_i}{\sigma^l_i}W^l_{ij}$ and $\tilde b^l_i=\frac{\gamma^l_i}{\sigma^l_i}\left(b^l_i-\mu^l_i\right)+\beta^l_i$. This makes it simple to convert BN layers into SNNs, because after transforming the weights of the preceding layer, no additional conversion for BN layers is necessary. Empirically we found loss-less conversion if BN parameters are integrated into other weights like this, the advantage lies purely in obtaining  better ANNs using BN during training.

\subsection{Analog input to first hidden layer}
\label{sec:analog_input}

%The ideal use of SNNs is to process event-based inputs, e.g. from DVS sensors \citep{lichtsteiner2008128}, but few benchmark datasets in this %representation exist \citep{hu2016dvs}, so 
Because truly event-based benchmark datasets are rare \citep{hu2016dvs}, conventional frame-based image databases such as MNIST \citep{lecun1998gradient} or CIFAR \citep{krizhevsky2009learning} have been used to evaluate the accuracy of the SNN after conversion. Previous work has usually converted analog input activations, e.g. gray levels or RGB values into Poisson firing rates. But this introduces variability into the firing of the network and impairs its performance, without having any notable benefits. A simple alternative is to use  analog input values in the very first hidden layer, and compute with spikes from there on \citep{zambrano2016fast}. Empirically we found this to be particularly effective in the low-activation regime of ANN units, where usually undersampling in spiking neurons poses a challenge for successful conversion.
% \BR{In their recent ArXiv publication, Zambrano and Bohte use analog input as well, without really explaining how/why. We obviously did not take our idea from them, so what do we do about this?}

\subsection{Spiking softmax}
Softmax is commonly used as the output of a deep ANN, because it results in normalized and strictly positive class likelihoods. Previous approaches for ANN-to-SNN conversion could not convert softmax layers, but simply predicted the class corresponding to the neuron that spiked most during the presentation of the stimulus. However, this approach fails when all neurons in the final layer receive negative inputs, and thus never spike. 
%In an ANN using the softmax operation this is no problem, because softmax would produce positive classification probabilities anyway.

Here we convert ANN softmax layers by using a mechanism proposed in \citep{nessler2009stdp}, where output spikes are triggered by an external Poisson generator with variable firing rate. The spiking neurons do not fire on their own but  simply accumulate their inputs. When the external generator determines that a spike should be produced, a softmax competition according to the accumulated membrane potentials is performed. 
%Instead, spikes of softmax neurons are triggered by an external Poisson-clock with parameter $\mu$. The value of $\mu$ is set between zero and the highest possible firing rate, given by the inverse time resolution $1/\mathrm{dt}$. If the Poisson-clock triggers a spike in the output layer, we still need to decide which neuron spikes. This is done by computing the softmax function over the membrane potentials of the neurons in the output layer. The neuron with the highest softmax probability fires a spike. The final classification result over the course of stimulus presentation is then given by the index of the neuron with the highest firing rate, as before.

\subsection{Spiking max-pooling layers}
Most successful ANNs use max-pooling to spatially down-sample feature maps, but this has not been used in SNNs because computing maxima with spiking neurons is non-trivial. Instead, simple average pooling had been used in \citep{cao2015spiking,Diehl2015Fast}, which results in using weaker ANNs before conversion. Lateral inhibition, as suggested in \citep{cao2015spiking}, does not fulfill the job properly, because it only selects the winner, but not the actual maximum firing rate. Another suggestion by \citep{orchard2015} is to use a time-to-first-spike encoding, in which the first neuron to fire is considered the maximally firing one. Here we propose a simple mechanism for spiking max-pooling, in which output units contain gating functions, which only let spikes from the maximally firing neuron pass, while discarding spikes from other neurons. The gating function is controlled by computing estimates of the pre-synaptic firing rates, e.g. by computing an online or exponentially weighted average. In practice we found several methods to work well, but demonstrate only results using exponentially weighted averages of firing rates to control the gating function.

\section{Results}
\label{sec:results}

There are two ways to improve the accuracy of the SNN via conversion: 1) training a better ANN before conversion, and 2) improving the conversion by eliminating approximation errors of the SNN. In the following we show the influence of both approaches.

\subsection{Contribution of improved ANN architectures}
\label{sec:annmethods}

The methods introduced in Section \ref{sec:methods} allow conversion of CNNs that use biases, softmax, batch-normalization, and max-pooling layers, which all improve the accuracy of the ANN. This was quantified on the CIFAR10 benchmark \citep{krizhevsky2009learning}, using a CNN with 4 convolution layers (32 3x3 - 32 3x3 - 64 3x3 - 64 3x3), ReLU activations, batch-normalization, 2x2 max-pooling layers after the 2nd and 4th convolutions, followed by 2 fully connected layers (512 and 10 neurons) and a softmax output. This ANN achieved 87.86\% accuracy. Constraining the biases to zero reduced the accuracy to 87.73\%. Replacing max-pooling by average-pooling further decreased the accuracy to 87.69\%. Eliminating the softmax and using only ReLUs in the output led to a big drop to 69.44\%. With our new methods we can therefore start the conversion already with much better ANNs than was previously possible.

%\begin{table}[t]
%\caption{Influence of network components on the accuracy of the underlying ANN. All networks use softmax in the output layer, except a), where the softmax is replaced by a ReLU activation. All networks use MaxPooling layers except b), where the Max is replaced by AveragePooling. All networks except c) use biases. Network d) uses all components, and achieves the best performance.}
%\label{tab:ANNtraining}
%\centering
%\begin{tabular}{ll}
%\toprule
%a) No softmax & 69.44 \\
%b) Avg instead of MaxPool & 87.69 \\
%c) No bias & 87.73 \\
%d) MaxPool, bias & 87.86 \\
%\bottomrule
%\end{tabular}
%\end{table}

\subsection{Contribution of improved SNN conversion methods}

\begin{figure}[htbp]
\begin{subfigure}{.4\textwidth}
\centering
\includegraphics[scale=0.4, trim={9cm 12cm 0 0}, clip]{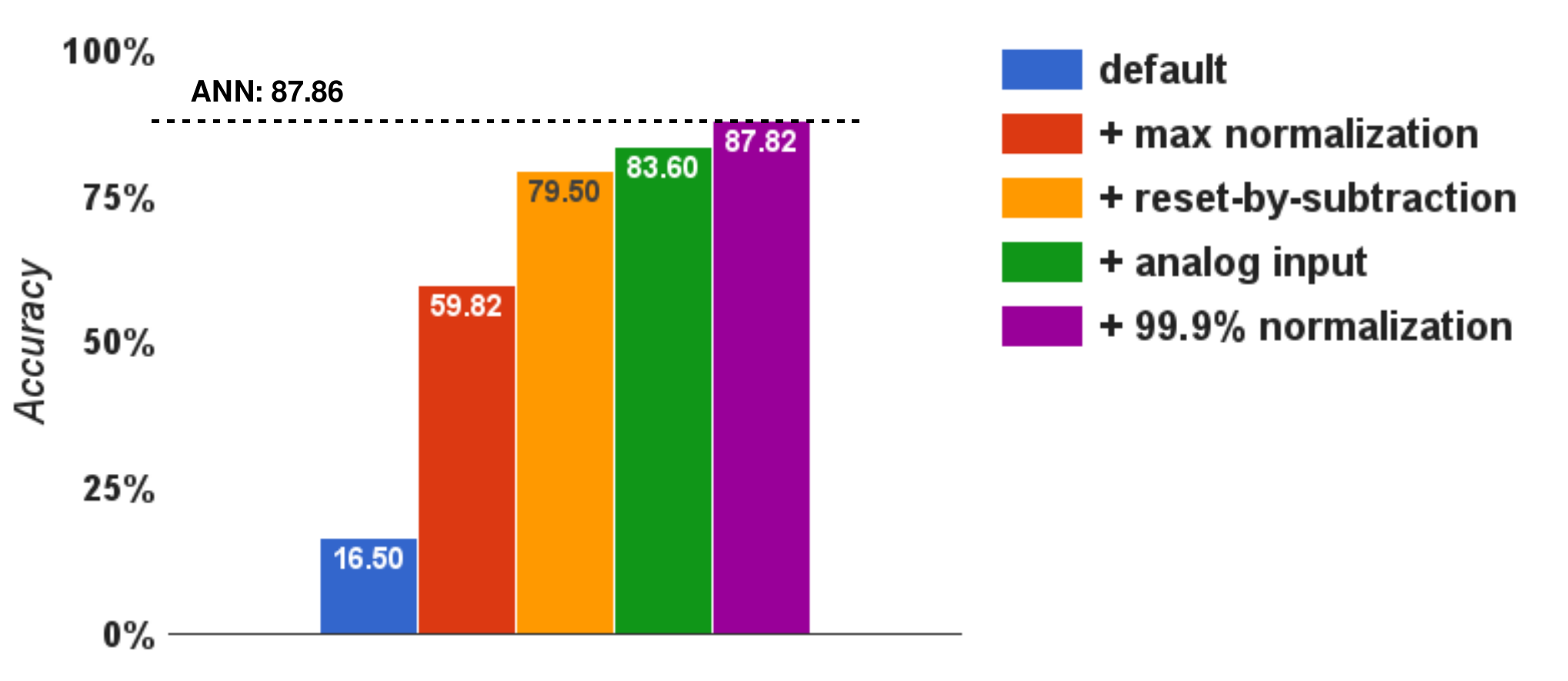}
\caption{}
\label{fig:SNNconversion}
\end{subfigure}
\hfill
\begin{subfigure}{.4\textwidth}
\centering
\includegraphics[scale=0.3]{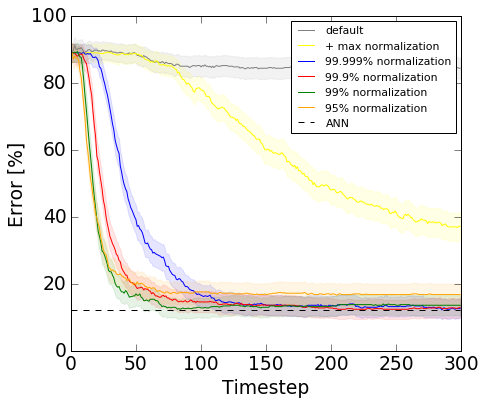}
\caption{}
\label{fig:error_vs_time}
\end{subfigure}
\caption{(a) Influence of novel mechanisms for ANN-to-SNN conversion on the SNN accuracy for CIFAR10. The best ANN from Section \ref{sec:annmethods} (87.86\%) is converted into an SNN. Default mode (blue bar): SNN with Poisson inputs, reset-to-zero, and no weight normalization. Red bar: applying weight normalization as in \citep{Diehl2015Fast}. For the next three bars we apply novel techniques as presented in Section \ref{sec:methods}. Shown is the accuracy after 300 time steps. (b) Accuracy-latency-tradeoff: SNNs give approximate results even when inputs are incomplete, and improve their accuracy with time. Tested on 400 CIFAR10 samples we find that the accuracy improves rapidly, and approaches the ANN level. The robust weight normalization factor can be tuned to achieve an ideal tradeoff between latency and final accuracy.}
\end{figure}

Figure \ref{fig:SNNconversion} shows that in the case of CIFAR10 the conversion of the best ANN into an SNN using the default approach (i.e. no normalization, Poisson spike train input, reset-to-zero) fails, yielding an accuracy of 16.5\%, barely above chance level. Adding weight normalization as suggested in \citep{Diehl2015Fast} (red bar) raises the accuracy to 59.82\%, but this is still a big drop from the ANN result of 87.86\%. Changing to the \emph{reset-by-subtraction} mechanism from Section \ref{sec:theory} leads to another 20\% improvement (orange bar), and switching to analog inputs to the first hidden layer instead of Poisson spike trains results in an accuracy of 83.6\% (green bar). Finally, using the 99.9th percentile of activations for robust weight normalization yields 87.62\% accuracy, which is very close to the ANN performance and our best result for CIFAR10 with single SNNs. We can therefore conclude that all the proposed mechanisms for ANN training and ANN-to-SNN conversion contribute positively to the success of the method. The conversion into a SNN is nearly loss-less, and the results are very competitive for classification benchmarks using SNNs. These results were confirmed also on MNIST, where a 7-layer network with max-pooling achieved an accuracy of 99.44\%, thereby improving previous state-of-the-art results for SNNs reported by \citep{Diehl2015Fast} and \citep{zambrano2016fast}. 

%On MNIST, the new methods presented here enabled us to convert an ANN with 99.44\% accuracy into an SNN with 99.44\% accuracy. This is substantially better than our previous record for SNNs reported in \citep{Diehl2015Fast}, as well as the recent achievement of 99.14\% by \citep{zambrano2016fast}. Units in our network typically spike at an average of 20 Hz, whereas their method needs only 8.6 Hz for the network performance to converge. However, the computations in their adaptive update and reset mechanism are more costly.

% Beyond MNIST and CIFAR10, tests on datasets such as Caltech101, SVHN and ImageNet are currently underway.

SNNs are known to exhibit a so-called accuracy-latency-tradeoff \citep{Diehl2015Fast,neil2016learning}, which means that the accuracy improves the longer the network is being simulated. In Figure \ref{fig:error_vs_time} we show that the robust weight normalization factor can be tuned to ideally exploit this property. Empirically the best results were obtained with normalization factors corresponding to the 99th or 99.9th percentiles of activations. Both lead to accurate classifications quickly, and also converge to error rates very similar to those of the underlying ANN.

\section{Discussion}
\label{sec:discussion}

By allowing a larger class of CNNs to be converted into SNNs, and by introducing a number of novel improved conversion techniques we could significantly improve the accuracy of our networks on both CIFAR10 and MNIST. Our best SNN result of 87.82\% accuracy on CIFAR10 compares favorably to previous SNN results: 
\citep{cao2015spiking} achieved 77.43\% accuracy on CIFAR10, albeit with a smaller network and after cropping images to 24x24. With a similarly small network and cropped images, \citep{Hunsberger2015Spiking} achieve 82.95\% accuracy. Both conversion methods lose less than 2\% due to the conversion, but since their approach does not use weight normalization, the results shown in Figure \ref{fig:SNNconversion} suggest that these methods would have problems on larger networks. Better SNN accuracies to date have only been reported by \citep{Esser20092016}, where an accuracy of 89.32\% was reported for a very large network optimized for 8 TrueNorth chips, and making use of ternary weights and multiple 1x1 network-in-network layers. A smaller network fitting on a single chip is reported to achieve 83.41\%. In our own recent experiments with similar low-precision training schemes for SNNs we converted the BinaryConnect model by \citep{Courbariaux2016}. Starting from an ANN with 91.94\% accuracy, we achieved an accuracy of the SNN of 91.35\% on CIFAR10, which is by far the best SNN result reported to date.

We know from ANNs that larger networks typically perform better than smaller networks, we thus fully expect to see a boost in SNN accuracy when applying the conversion techniques to even larger networks. In fact, the best ANNs to date achieve less than 5\% error on CIFAR10 \citep{springenberg2014striving}. Our goal here was to expand the toolkit for ANN-to-SNN conversion to the point where such large networks, using typical CNN mechanisms, can be converted into SNNs with only minimal loss of accuracy. That the drop-off is typically less than 1\% is encouraging. We have also shown that the typical accuracy-latency tradeoff is still present, although our networks were not specifically trained to converge fast. Using the techniques proposed by \citep{neil2016learning} should yield accurate results even faster.

%\subsection{Neurons with bias}
%Biases are routinely used in all common ANN architectures. In SNNs biases are typically avoided, because spiking neuron models are often grounded in computational neuroscience, and constant firing is considered biologically implausible, in addition to violating the desired purely data-driven mode of inference. However, typical biases in deep networks tend to be small and equally likely positive or negative, so there is no big increase of firing in a network that uses biases. Also, because the biases are constant, it is possible to implement a completely event-based update of neurons even in the presence of biases, because the next expected spike time can be easily calculated (or updated when input spikes arrive).

%\subsection{Spiking softmax}
%This spiking softmax implementation allowed us to further reduce the gap in accuracy after ANN-to-SNN conversion. In some networks the effect was insignificant because the naive max-rate version already performed optimally. However, when all output activations are negative, which was the case for one network we trained, the SNN performance dropped to chance level because the output layer was completely silent.

\section{Conclusions}
\label{sec:conclusions}

Deriving a first solid theory for ANN-to-SNN conversion has directly revealed mechanisms to improve the classification accuracy of the resulting SNN by a simple switch of reset mechanisms.
%In particular, \citep{zambrano2016fast} have applied a reset-by-subtraction mechanism without providing an analytical argument for it.  We showed that reset-by-subtraction preserves the information contained in the overshoot in membrane potential above threshold, which otherwise is discarded in standard reset-to-zero.
This, together with novel tools to convert a large class of CNNs, covering most standard features of conventional CNNs, has helped achieving state-of-the-art SNN results, and almost loss-less ANN-to-SNN conversion. Future research will apply the methods to new datasets that require large networks, such as ImageNet. Another promising line of research is to investigate mechanisms that further reduce the number of spikes produced, eliminating redundancies when information about static inputs are sent. We have demonstrated that the accuracy gap between ANNs and SNNs can be almost completely closed.

\subsubsection*{Acknowledgments}

We thank Jun Haeng Lee for helpful comments and discussions. This work has been supported by the Samsung Advanced Institute of Technology.
% Do not include acknowledgments in the anonymized submission, only in the final paper.

\small

\bibliographystyle{apalike}
\bibliography{Mendeley,newbib}

\end{document}